# Ultra-Fast Zernike Moments using FFT and GPU

Mohammed Al-Rawi,
DETI, IEETA, University of Aveiro,
Aveiro, Portugal
al-rawi@ua.pt

**Abstract**

Zernike moments can be used to generate invariant features that are applied in various machine vision applications. They, however, suffer from slow implementation and numerical stability problems. We propose a novel method for computing Zernike using Fast Fourier Transform (FFT) and GPU computing. The method can be used to generate accurate moments up to high orders, and can compute Zernike moments of 4K resolution images in real-time. Numerical accuracies of Zernike moments computed with the proposed FFT approach have been analyzed using the orthogonality property and the results show that they beat other methods in numerical stability. The proposed method is simple and fast and can make use of the huge GPU-FFT libraries that are available in several programming frameworks.

## 1 INTRODUCTION

Zernike moments (ZMs) can be used to derive useful invariant features and have found many applications in various image related research [1, 2]. The history of ZMs goes back to Teague [3] who introduced the notion of the orthogonal moments using a kernel composed of Zernike Radial Polynomials (ZRPs) and Euler function (EFs), the latter is also called angular Fourier complex function. Using Jacobi polynomials, however, Bhatia and Wolf pointed out that there is an infinite number of complete sets of orthogonal radial polynomials [4]. An interesting feature of orthogonal moments is that they can be used to derive features that are invariant to translation, rotation, and scale. The orthogonality property enables the reconstruction of the image from the computed moments, the reconstructed image is a standard orthogonal series estimate defined by a truncation parameter representing the maximum order. Other types of orthogonal moments have also been introduced recently based on different orthogonal radial polynomials, but using the same EFs for the angular part. The computation of 2D Zernike Moments using the direct computation of Zernike radial polynomials (ZRPs) is not only slow, but results in huge errors due to the numerical instability of ZRPs [5, 6]. This led to several other methods that provided a significant step towards reducing the computations of orthogonal moments, as in [7]. Whilst other works focused on reducing the computational errors endued in the computation and on investigating the numerical stability, as in [5, 6]. The extension of the moment methods to 3D images and 3D data has also been of interest and can be used in 3D object recognition and detection [8, 9]. The methods have been applied in several applications related to pattern recognition and computer vision [10].

Unfortunately, the computation of ZMs is hindered by the operations needed to calculate the ZRPs. Several methods have been proposed to recursively generate ZRPs and thus reduce their computational complexity. The so-called *q*-recursive method has been proposed to reduce the computations [11]. Kintner's method has also been proposed to resolve the ZRPs computational complexity problem, but it has the drawback that it is based on ZRPs order recursion (thus the name p-recursion comes), thus, the whole ZMs cannot be computed for one order unless computing all the orders [11]. To sum, these methods, as well as the direct implementation, still suffer from numerical instability when higher order moments are considered.

A method that can be used to efficiently compute ZRPs, by exploiting their relation to the Fast Fourier Transform (FFT), has been proposed in [12]. This method, however, has not been used in image processing nor in the 2D Zernike moment pattern recognition literature. The method presented in [12], which was only intended to compute ZRPs and not ZMs, aimed at investigating the aberrations of optical systems. Hence, this could be the main reason for not finding its way into image processing and pattern recognition applications. One way to utilize the method is by utilizing it to efficiently compute ZMs, which to the best of our knowledge, is an approach that has not been tackled before. In this work we propose making use of ZRPs computed via FFT and GPU computation to efficiently and accurately calculate ZMs. Exploring the orders of speed and accuracy improvements that this method can achieve is of high interest to researchers and developers working on image analysis and pattern recognition. To reduce the image reconstruction error, we also introduce Neumann factor normalization for ZMs. Several ZMs-FFT image reconstruction experiments have been performed, with and without the Neumann factor, to demonstrate the speed gain, numerical accuracy, and how much this normalization adds to ZMs.

## 2 THEORY AND METHODS

Orthonormal moments outperform geometric moments in capturing the image characteristics, and thus, better object recognition can be achieved. Orthogonal moments use orthogonal polynomials as basis functions. When ZRPs and EFs are used as bases functions, the resultant coefficients are called Zernike moments (ZMs). Moreover, 2D and 3D ZMs can be used for 2D and 3D object recognition/analysis, respectively. This work will focus on 2D ZMs, simply denoted as ZMs, defined as follows:

$$Z_{nm} = \frac{n+1}{\pi} \iint_{0 \leq \rho \leq 1,\ 0 \leq \theta \leq 2\pi} \hat{f}(\rho, \theta) V_{nm}(\theta)^* \rho d\rho d\theta, \qquad (1)$$

where $Z_{nm}$ denotes a ZM of order $n$ and repetition $m$, $V_{nm}(\rho, \theta)^* = R_{nm}(\rho)e^{-jm\theta}$ is usually called Zernike function, $j = \sqrt{-1}$, $\hat{f}$ is the polar form of the image $f$ that is typically acquired in Cartesian coordinates, * denotes the complex conjugate and $R_{nm}(\rho)$ denotes ZRPs of order $n$ and repetition $m$. The order takes on positive integer values $n \geq 0$, and the repetition $m$ takes on positive and negative integer values subject to the conditions $n - m$ is even and $|m| \leq n$. ZMs with negative $m$ values can be found from $Z_{n,-m} = Z_{nm}^*$. ZRPs are defined:

$$R_{nm}(\rho) = \sum_{s=0}^{(n-|m|)/2} \frac{(n-s)!(-1)^s}{\left(\frac{n-|m|}{2}-s\right)!\left(\frac{n+|m|}{2}-s\right)!s!}\rho^{n-2s},\qquad(2)$$

such that $0 \leq \rho \leq 1$, where $\rho$ is the radial parameter inside the image. The computation of ZMs has two hindering computational processes, the first is the generation of ZRPs using (2), and the second is the accumulation of ZMs using (1). The computation of ZRPs could be perturbed due to using fixed-point arithmetic; hence in practice, the values of $R_{nm}$, especially at high orders, may degrade, leading thus to numerical instability. The computation of (1) is usually performed using numerical integration that requires interpolating the image to the polar form. Therefore, it is more convenient to write ZMs via Cartesian summation in the Cartesian coordinate system, as follows:

$$\tilde{Z}_{nm} = \frac{n+1}{\pi} \sum_x \sum_y v^*_{nm}(x,y) f(x,y), \qquad(3)$$

where

$$v^*_{nm}(x,y) = \int_{x-\Delta/2}^{x+\Delta/2} \int_{y-\Delta/2}^{y+\Delta/2} V_{nm}(x,y)^* \, dx\, dy, \qquad(4)$$

$\Delta = 2/M$ is the pixel width, and the value of $v^*_{nm}(x,y)$ should be computed using numerical integration. One, however, can use first order approximation for $v^*_{nm}(x,y)$, although this is the worst case error scenario, but less computations are needed. This approximation can be written as follows:

$$v^*_{nm}(x,y) \approx \Delta^2 V_{nm}(x,y)^*. \qquad(5)$$

Clearly, the discrete ZM can be rewritten as:

$$Z_{nm} = \frac{(n+1)}{\pi N^2} \sum_x \sum_y R_{nm}(\rho) f(x,y) e^{-jm\theta}, \qquad(6)$$

where $\rho = \sqrt{x^2 + y^2}$, and $\theta = y/x$. It is worth mentioning that the error in ZMs computation is not only related to the numerical integration of (4) and the digitization errors of (6), but also to the numerical stability in computing ZRPs. If the computation is done using (1), then there is an error provoked by the mapping of the rectangular image grid to a circular one, an error so often referred to as the geometric error. Nonetheless, two major approaches can be used together to efficiently compute Zernike moments i) by making use of the symmetrical properties of the exponential term $e^{-jm\theta}$, ii) and by reducing the computations of the radial polynomial term $R_{nm}(\rho)$. Further details on the computation of ZMs are shown below. The computational complexity of ZMs thus depends heavily on the computation of $R_{nm}(\rho)$. Fortunately, several algorithms have been developed to facilitate the computation of $R_{nm}(\rho)$, for example, the p-recursive method, Prata's method, Kintner's method, and the $q$-recursive.

## 2.1 Fast Computation of Zernike Radial Polynomials via FFT

It has been shown in [12] that ZRPs has the following discrete Fourier transform:

$$R_{nm}(\rho) = \frac{1}{N} \sum_{k=0}^{N-1} U_n\left(\rho \cos\cos \frac{2\pi k}{N}\right) \cos\cos \frac{2\pi mk}{N}, \quad (7)$$

where the value of $N$ is at least $N = 2n + 1$, and $U_n$ is the Chebyshev polynomial of the second kind and of degree $n$ that can be found using,

$$U_n(x) = \frac{\sin\sin\,(n+1)v}{\sin\sin v}, \quad x = \cos\cos v. \quad (8)$$

Obviously, equation (7) has the form of a discrete cosine transform (i.e., real part of FFT).

## 2.2 Image Reconstruction from Zernike Moments

Image reconstruction is an important concept that can be used to test how well, and to which order, ZMs emulate the original image that has been used to calculate them. This method is also known as the moment inverse problem. That said, after finding ZMs, one can reconstruct the original image using,

$$\widetilde{f}(x,y) = Re\left[\sum_{n=0}^{n_{max}} \sum_{\substack{m=-n \\ n-m,\ is\ even}}^{n} Z_{nm} R_{nm}(\rho_{xy}) e^{jm\theta_{xy}}\right], \quad (9)$$

where $n_{max}$ is the maximum order one may choose to compute ZMs, i.e. the set $\{Z_{nm}: n = 0, 1, ..., n_{max}; m = -n, -n+2, ..., n\}$. $\rho_{xy} = \sqrt{x^2 + y^2}$, $\theta_{xy} = (y/x)$, $Re[\zeta]$ refers to taking the real part of $\zeta$. As can be seen from (9), image reconstruction from Zernike moments, which is also referred to as "the moment inverse problem", requires heavy computations. For complete and exact reconstruction, however, one should work with $n_{max} \to \infty$, but one essentially needs to work with finite orders.

## 2.3 Zernike Moments of Colour Images

A color image is represented by three bands, red, green and blue, which can also be called an RGB image. To compute ZMs for a color image, each color band is processed separately and the minimum and maximum values for each band are stored along with ZMs(R), ZMs(G), and ZMs(B). To reconstruct the color image, the reconstructed image of each band is then normalized to have the minimum and maximum values of the original band.

## 2.4 The Reconstruction Error

To measure the performance of the reconstructed image and the ZM method used, an error measure is used to compare the reconstructed image to the original image. Image reconstruction error is usually calculated according to:

$$\epsilon_1 = \frac{\sum_x \sum_y (f(x,y) - \tilde{f}(x,y))^2}{\sum_x \sum_y f(x,y)^2}, \quad \forall\ x^2 + y^2 \leq 1. \tag{10}$$

The condition $x^2 + y^2 \leq 1$ in the error measure shown above is necessary to ensure excluding the zeros outside the unit circle bounding the image. However, the above formula is not accurate enough to judge the quality of the reconstructed image. For example, it is possible that the above formula gives $\epsilon > 1$ when

$$f(x,y) \cup \tilde{f}(x,y) = \emptyset, \quad \forall\ x^2 + y^2 \leq 1, \tag{11}$$

which may lead to having $\sum_x \sum_y (f(x,y) - \tilde{f}(x,y))^2 > \sum_x \sum_y f(x,y)^2$. Another error measure also appeared in [13] as follows:

$$\epsilon_2 = \sum_x \sum_y \frac{(f(x,y) - \tilde{f}(x,y))^2}{f(x,y)^2}, \quad \forall\ x^2 + y^2 \leq 1, \tag{12}$$

which suffers from singularity for any point at which $f(x,y) = 0$. Therefore, a better alternative than the two above error measures is the following:

$$\epsilon = \frac{\sum_x \sum_y (f(x,y) - \tilde{f}(x,y))^2}{f_{max}^2 \sum_x \sum_y}, \quad \forall\ x^2 + y^2 \leq 1. \tag{13}$$

where $f_{max}$ denotes the maximum gray-level value in $f$. If needed, the peak signal to noise ratio can be obtained as $PSNR = \sqrt{\epsilon}$. In color images, the reconstruction error is calculated separately for each band.

## 2.5 The Interpolation Dilemma in Finding $\hat{f}(\rho, \theta)$ and the Numerical Integration

Digital images are usually acquired in a Cartesian coordinate system, and they are generally arranged into a rectangular grid. Unfortunately, ZMs are represented in the polar domain. The computation of a polar integration can be performed by uniformly changing the values of $0 < \rho < 1$ and $0 \leq \theta \leq 2\pi$ and then interpolating the value of $\hat{f}(\rho, \theta)$, which may result in interpolation errors. Instead of this computation scheme that requires to map $f$ from $(x, y)$ to $(\rho, \theta)$, it is more appropriate to do the computation using $f$ in Cartesian coordinates and one can map only the coordinate values from $(x, y)$ to $(\rho_{xy}, \theta_{xy})$ and then use the set $\{(\rho_{xy}, \theta_{xy}): x = 1, ..., N; y = 1, ..., N;$ such that $\rho_{xy} \leq 1\}$ to calculate Zernike functions.

Clearly, the benefit of working in the Cartesian coordinate system is to reduce the interpolation errors and its heavy computations. Nonetheless, less exhaustive computations can be performed if one considers using polar coordinates and image interpolation, and implementing more advanced numerical integration methods such as adaptive Gauss-Kronrod quadrature.

### 2.6 Neumann Factor Normalization

To reduce the error induced by embedding the image inside the unit circle (which is a necessary condition to compute ZMs), we propose to normalize ZMs by dividing them by the Neumann factor, as follows:

$$Z_{nm} = \frac{(n+1)}{\epsilon_m \pi N^2} \sum_x \sum_y R_{nm}(\rho) f(x,y) e^{-jm\theta}, \quad (14)$$

where $\epsilon_m$ is the Neumann factor (called so because it frequently appears in conjunction with Bessel functions), which is given by:

$$\epsilon_m = [2 \; if \; m = 0 \;\; 1 \; otherwise]. \quad (15)$$

### 2.7 Using Zernike Moments to Clean Computer Vision Datasets

It is possible that large computer vision datasets to contain repeated images, and to achieve better analysis and results, it is recommended to remove such redundancy. This efficient approach, in fact, can be used to the proximity of the different images in any image dataset, including videos. In this section, we propose a novel algorithm that can be used to remove redundant images in computer vision dataset. The algorithm depends on finding the intersection of several ZM vectors that are extracted from each image. ZMs will thus be used as a hashing function, namely $f$, denoted as:

$$f_l = ZM_l(f_k), \quad (16)$$

where $l = 1, 2, ...L$ is the number of ZM order that will be used, and where $k = 1, 2, ...K$ is the number of images in the dataset. The intersection for orders $l_1$ and $l_2$ thus yields:

$$f_{l_1} \cap f_{l_2} => i_{l_1,l_2}, \quad (17)$$

$i_{l_1,l_2}$ is a set that contains the indices of images that have exactly the same content.

## 3 EXPERIMENTAL RESULTS

We used various schemes to investigate the reconstruction of images up to extremely high orders. In all experiments, we used a 256×256 Lena RGB image. There are a few steps that one needs to consider in ZMs computation, such as:

- Normalizing the gray-level values of the reconstructed image to [min, max], where min and max are the minimum and the maximum gray-level values of the original (input) image; for color images they are the minimum and the maximum values for each color band. The power of this normalization is demonstrated in Table I.

- Prior to the computation of ZMs, the original $N \times N$ image was embedded in a larger grid (zero-valued image) that has the size $N + \lfloor N(\sqrt{2} - 1) \rfloor + 20$. Thus, the zero-valued image that will be used will have the size 383×383, which will host Lena (256×256) its central region. Hence, the offset value will be given by $\{\lfloor N(\sqrt{2} - 1)/2 \rfloor + 10, \lfloor N(\sqrt{2} - 1)/2 \rfloor + 10\}$.

TABLE I. IMAGE RECONSTRUCTION WITH ZERNIKE MOMENTS USING FFT AND NEUMANN FACTOR UP TO ORDER N=200, A 5×5 SEGMENT WAS EXTRACTED FROM EACH IMAGE AT THE REGION (126 TO 130) × (126 TO 130). THE PURPOSE OF THIS TABLE IS TO SHOW THE EFFECT OF MIN MAX NORMALIZATION ON THE RECONSTRUCTED IMAGE.

|  | Red band | Green band | Blue band |
|---|---|---|---|
| Original image | 253 253 254 246 232<br>253 252 248 248 243<br>251 253 245 244 236<br>253 249 250 240 236<br>253 244 248 235 239 | 215 202 188 172 151<br>210 203 175 165 160<br>208 197 177 173 136<br>209 189 182 151 152<br>208 198 172 147 147 | 166 151 127 118 127<br>161 140 124 114  97<br>159 134 134 114 102<br>150 141 122 111 104<br>141 157 114 131 105 |
| Without normalizing the reconstructed image | 7052 7112 7203 7267 7250<br>6978 6835 6737 6692 6662<br>6539 6139 5841 5689 5631<br>5649 5039 4631 4455 4401<br>4437 3760 3405 3339 3352 | 6260 6183 5713 4931 4058<br>5745 5148 4303 3416 2710<br>4641 3617 2590 1805 1390<br>3085 1871  936  462  398<br>1388  285    0    0    0 | 5789 5606 5105 4401 3698<br>5171 4590 3850 3146 2657<br>4203 3350 2543 1989 1777<br>3067 2158 1491 1209 1266<br>1985 1247  885  932 1194 |
| The reconstructed image normalized according to the min and max values (in each color band) of the original/input image | 202 203 205 206 206<br>200 198 196 195 194<br>192 184 179 176 175<br>175 163 156 152 151<br>152 139 132 131 131 | 173 171 161 145 126<br>162 149 131 113  98<br>138 117  95  79  70<br>106  80  60  50  49<br> 70  46  34  34  39 | 163 160 150 136 122<br>151 139 125 111 101<br>132 115  99  88  84<br>109  91  78  73  74<br> 88  73  66  67  72 |

## 3.1 Numerical Stability

To investigate numerical stability in ZMs, the proposed method is compared with the *q*-recursive method, as shown by Fig. 1. It is clear that the *q*-recursive method has numerical instability problems with orders higher than 150, since the reconstruction error increases with the order. This is due to the propagated arithmetic error, which becomes clear at high orders. However, the proposed FFT method has stable accuracy even up to order 500. Even at lower orders where the *q*-recursive has shown 0.08 reconstruction error compared to 0.11 using FFT, the reconstructed image using FFT has better subjective quality than the one reconstructed via the *q*-recursive.

## 3.2 Image Reconstruction Error

We compared the image reconstruction error of ZMs computed via FFT to that of the *q*-recursive method (regarded as one of the best methods in the literature) and the results are depicted in Fig. 2. The results of selected reconstructed images are shown in Fig. 3. As depicted in Fig. 2-c, Neumann factor normalization has led to reduced reconstruction error, and better subjective image quality as illustrated in Fig. 3. It is obvious that the ZMs via FFT outperforms the *q*-recursive method. However, the reconstructed images are a form of distortion, that we call the torchlight effect, and which we solved by proposing to normalized ZMs with Neumann factor.

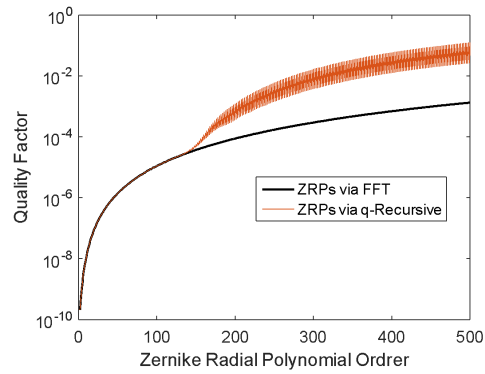

Fig. 1. Numerical stability Quality Factor (QF) that is based on orthogonality of Zernike Radial Polynomials -ZRPs [5]. The comparison of ZRPs computed via FFT and *q*-recursive. The lower the QF value the better accuracy the method gives. QF values are bonded between 0 and 1.

In Fig. 4, we present the famous Lena image reconstructed from ZMs up to order 480 and corrected with the Neumann factor (as can be seen, the reconstructed image is highly similar to the original). To demonstrate the speed improvement, we present in Fig. 5 several tests of the ZM-FFT computation, where 8-point symmetry [14] has also been considered to speedup the computations.

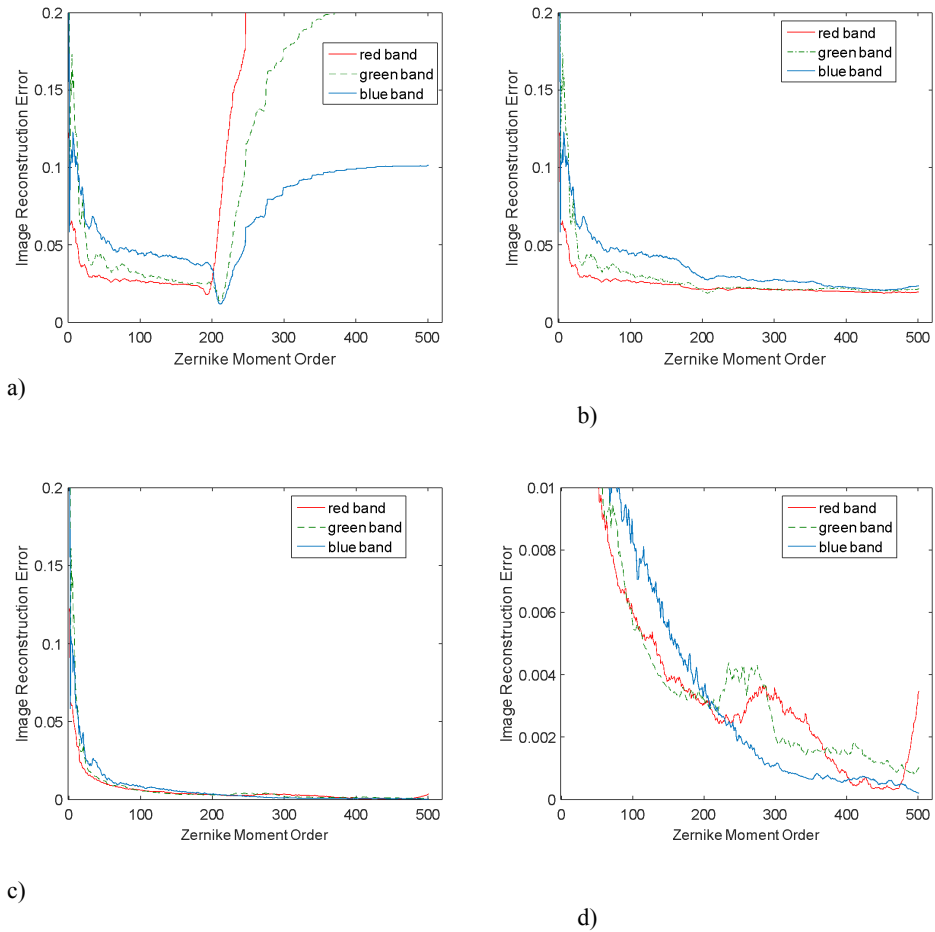

Fig. 2. Reconstruction of Lena image with Zernike moments a) *q*-recursive method, b) FFT method without Neumann factor correction, c) FFT method with Neumann factor correction, d) Enlarged view for E<0.01 of the error plot shown in (c).

## 4 CONCLUSION

The proposed computation of Zernike moments using fast Fourier transform is more accurate than the *q*-recursive method. Furthermore, the method is resilient and numerically stable up to extremely high orders. After performing tens of thousands of computational tests using Zernike moments and their related image reconstruction, we discovered that dividing Zernike moments by the Neumann factor reduces the image reconstruction error as well as the subjective quality of the image. The reconstruction thus of a color image from Zernike moments of that color image can be done accurately. Nonetheless, the min and max values of each color band of the original image need to be stored along with the values of Zernike moments of each band in order to use them in the normalization. In fact, the Lena (color) image reconstructed with Zernike via FFT and Neumann factor normalization is completely comparable to the original. This enhanced image reconstruction will lead to enhanced Zernike moment watermarking, as well as the generation of invariants that are used usually in computer vision and image analysis applications. In addition, the proposed method is extremely fast and can be used in real-time analysis of 4K high-resolution images.

| ZM order | 50 | 100 | 200 | 300 | 400 | 500 |
|---|---|---|---|---|---|---|
| *q*-recursive | <br> |  |  |  |  |  |
| ZMs-FFT, without Neumann Correction | <br> |  |  |  |  |  |
| ZMs-FFT, with Neumann Correction | <br> |  |  |  |  |  |

Fig. 3. Zernike moments reconstruction analysis of Lena image using the q-recursive method, FFT method without Neumann factor correction, and FFT method with Neumann factor correction

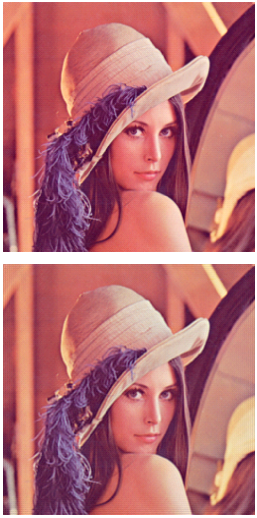

Fig. 4. Color image reconstructed up to order *n*=480 Zernike Moments, FFT algorithm and Neumann factor correction (right) compared to the original Lena (left).

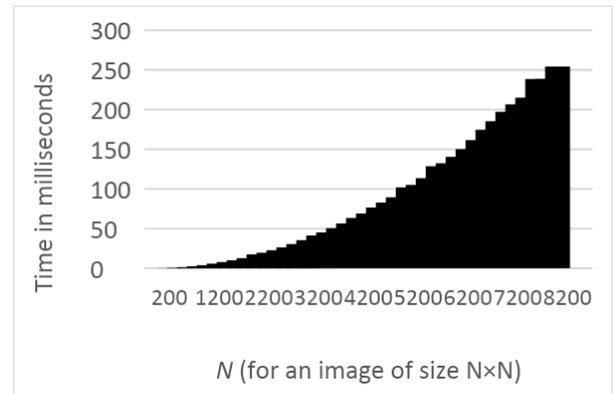

Fig. 5. Computation time in milliseconds, of any order single Zernike moment, with respect to image size (N), using FFT and GPU computing. One can calculate up to 20 FPS when the image size is 4000×4000. A single Nvidia Titanx xp GPU has been used in the analysis. The implementation time measurements for each N are an average of 1000 trials.